\documentclass{article}

\usepackage{microtype}
\usepackage{graphicx}
\usepackage{subcaption}
\usepackage{booktabs}
\usepackage{hyperref}

\usepackage[accepted]{icml2026}

\usepackage{amsmath}
\usepackage{amssymb}
\usepackage{mathtools}
\usepackage{amsthm}
\usepackage{algorithm}
\usepackage{algorithmic}

\usepackage[capitalize,noabbrev]{cleveref}

\theoremstyle{plain}

\theoremstyle{definition}

\theoremstyle{remark}

\icmltitlerunning{Evolutionary Curriculum Learning for Biological Sequence Modeling}

\begin{document}

\twocolumn[
  \icmltitle{Evolutionary Curriculum Learning Improves\\Biological Sequence Modeling}

  \icmlsetsymbol{equal}{*}

  \begin{icmlauthorlist}
    \icmlauthor{Richard Zhu}{cs,stat}
    \icmlauthor{Kento Nishi}{cs}
  \end{icmlauthorlist}

  \icmlaffiliation{cs}{Department of Computer Science, Harvard University, Cambridge, MA, USA}
  \icmlaffiliation{stat}{Department of Statistics, Harvard University, Cambridge, MA, USA}

  \icmlcorrespondingauthor{}{rzhu@college.harvard.edu, kentonishi@college.harvard.edu}

  \icmlkeywords{curriculum learning, variational autoencoder, biological sequences,
    protein variant effect prediction, RNA design, multiple sequence alignment}

  \vskip 0.3in
]

\printAffiliationsAndNotice{}

\begin{abstract}
Variational autoencoders (VAEs) trained on multiple sequence alignments (MSAs) have
emerged as powerful generative models for biological sequences, with applications
ranging from disease variant prediction to functional RNA design.
However, standard biological VAE training treats all sequences as exchangeable, ignoring
the rich evolutionary structure that organizes homologous sequences from evolutionarily
close to highly divergent.
We propose \emph{Evolutionary Curriculum Learning} (ECL), a training strategy that
exploits this structure by progressively exposing the model to sequences of increasing
evolutionary distance from sampled anchors, following a power-law expansion schedule.
Applied to two architecturally distinct VAE models and two biological domains—protein
variant effect prediction with EVE and RNA family sequence generation with
RfamGen—ECL improves downstream task performance across five random seeds per
configuration.
Mean ClinVar classification AUROC rises from $0.981$ to $0.989$ for p53; for PTEN,
ECL attains $1.000$ in every seed whereas the baseline is unstable (mean $0.905$,
falling as low as $0.54$).
For RNA, ECL raises mean covariance-model bit scores on all three families tested and
exceeds its seed-matched baseline in 12 of 15 training runs, though with only three
families the effect cannot be established as significant at the family level.
Ablation experiments show that progressively expanding the sampled sequences by evolutionary distance outperforms fixed-size neighborhood sampling in addition to uniform random sampling. Evolutionary distance is therefore a useful inductive bias for ordering the training
curriculum in biological sequence modeling.
\end{abstract}

\section{Introduction}
\label{sec:intro}

The evolutionary record encoded in multiple sequence alignments (MSAs) is the primary
signal exploited by alignment-based generative models of biological sequences.
By learning the distribution of naturally occurring sequence variation, VAEs trained on
MSAs implicitly capture the structural and functional constraints that govern a homologous protein set or RNA family.
This paradigm has enabled remarkable advances: EVE~\citep{frazer2021disease} predicts
the pathogenicity of missense variants at a level competitive with high-throughput
functional assays, while RfamGen~\citep{sumi2024deep} designs novel functional RNA
sequences by sampling from a learned latent space structured around RNA sequence and structure.

Despite these successes, a fundamental property of MSAs has been overlooked during
training: sequences in an MSA are not exchangeable and should not be randomly sampled
to form batches during training. Rather, they span a continuous spectrum of
evolutionary distances, with some sequences separated by only a few mutations while
others have diverged substantially over time. This heterogeneity creates a natural
hierarchy of learning difficulty: evolutionarily close sequences tend to provide more
locally consistent structural and functional constraints, whereas distant sequences
introduce greater compositional and structural variation that may be harder to model.
Standard stochastic gradient training does not exploit this structure when forming
minibatches: it either samples nearby and distant sequences with equal probability or
uses phylogenetic reweighting to downweight sequences that have many close neighbors
in the MSA, as in both \citet{frazer2021disease} and \citet{sumi2024deep}.

Curriculum learning~\citep{bengio2009curriculum} leverages this hierarchical structure by
organizing training examples from easy to hard.
Applying this principle to MSA-based generative models is natural: evolutionary
distance is a biologically-principled, data-derived proxy for difficulty that
requires no additional annotations or learned scoring.

We introduce \textbf{Evolutionary Curriculum Learning (ECL)}, a plug-in training
strategy for MSA-based VAEs. Instead of sampling training sequences uniformly,
ECL samples a randomly chosen anchor and then draws minibatches from an
expanding neighborhood of evolutionarily similar sequences.
Because ECL only changes the sampling order, it requires no architectural
modifications, only pairwise distances computed once during preprocessing.

\begin{figure}[t]
  \vskip 0.1in
  \begin{center}
    \centerline{\includegraphics[width=\columnwidth]{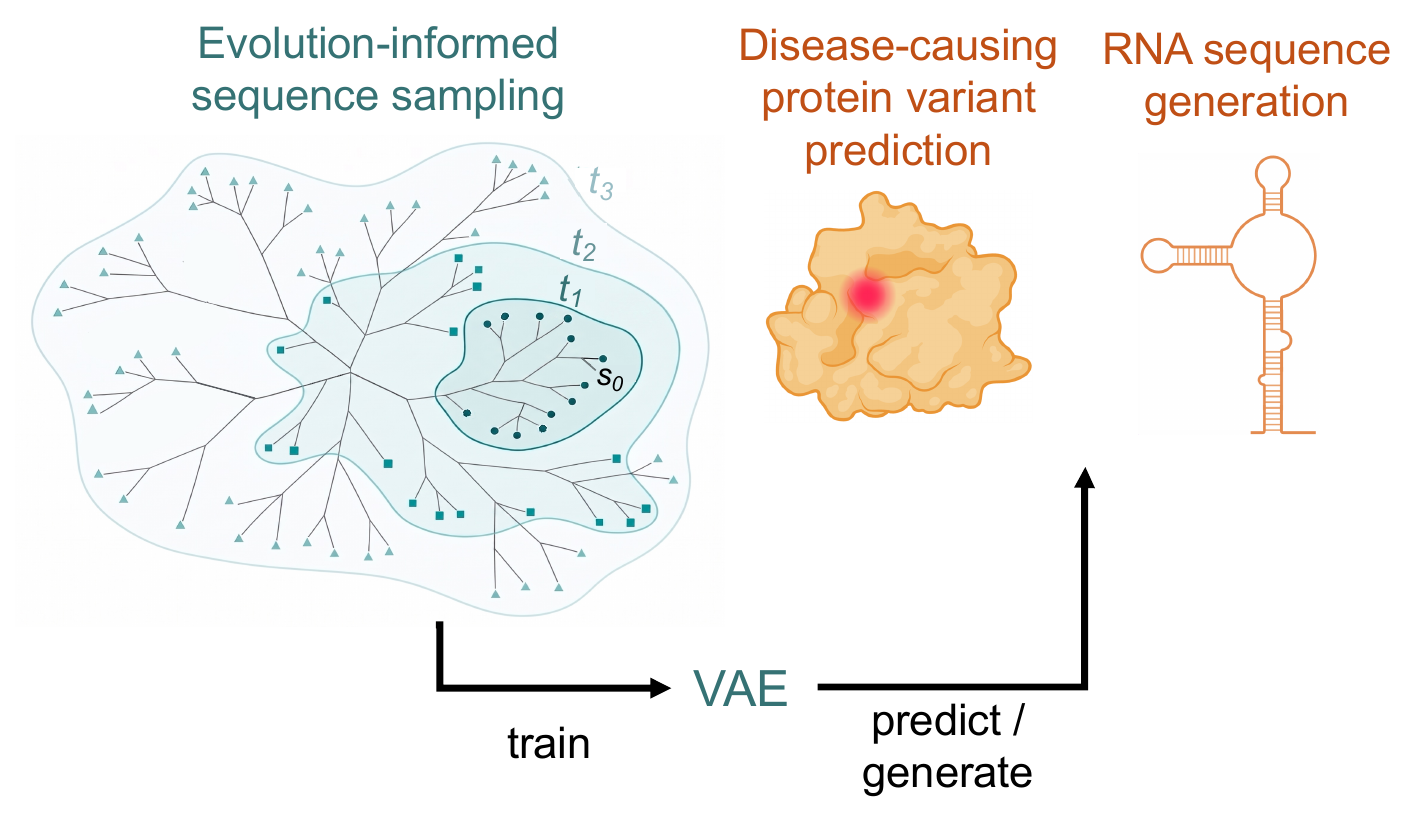}}
    \caption{
      Overview of evolutionary curriculum learning: over multiple time steps $t_i$, increasingly evolutionarily distant neighbors of sampled anchors are added to the set from which sequences are sampled to train the VAE model for modeling sequence variation. We demonstrate the ability for this framework to improve prediction of disease-causing protein variants and RNA family sequence generation.
    }
    \label{fig:ecl_overview}
  \end{center}
  \vskip -0.1in
\end{figure}

Figure~\ref{fig:ecl_overview} illustrates the idea. In the rest of the paper, we
show that this simple curriculum learning technique improves performance of two VAE-based biological sequence models: EVE for protein variant effect prediction and RfamGen for RNA family sequence generation.

\section{Background}
\label{sec:background}

\paragraph{EVE.}
EVE~\citep{frazer2021disease} trains a Bayesian VAE on a deep MSA of homologous
proteins retrieved from UniRef.
After training, the pathogenicity of a missense variant $v$ is scored via the
evolutionary index, the negative change in estimated ELBO relative to the wild type.
A global Gaussian mixture model maps evolutionary indices to pathogenic/benign/uncertain
class assignments.
EVE achieves a mean AUROC of 0.91 for ClinVar pathogenic variant classification across 3,219 proteins—on par with or
better than experiments.

\paragraph{RfamGen.}
RfamGen~\citep{sumi2024deep} integrates covariance model (CM) grammar structure into a VAE.
A covariance model~\citep{eddy1994rna} is a probabilistic grammar that jointly represents an RNA multiple sequence alignment and its consensus secondary structure, so it can capture both sequence conservation and base-pairing constraints.
In RfamGen, each aligned sequence is parsed into a triplet
$(tr, ss, bp)$ of one-hot matrices representing CM transition rules, single-strand
emissions, and base-pair emissions, respectively.
The encoder and decoder operate over this structured representation.
Generated sequence quality is evaluated with CM bit scores---log-likelihood scores
assigned by the Rfam ground-truth CM---where higher values indicate sequences that
better resemble natural RNA family members.

\paragraph{Curriculum learning.}
\citet{bengio2009curriculum} showed that organizing training examples from easy to
hard improves generalization and convergence.
In biological sequences, evolutionary distance is an unsupervised, theory-grounded
difficulty proxy: conservation across close homologs reflects locally consistent functional and structural
constraints, while distant homologs introduce greater variation.

\section{Evolutionary Curriculum Learning}
\label{sec:method}

\subsection{Evolutionary Distance}

Given an MSA of $N$ sequences $\{s_i\}_{i=1}^N$, we compute all pairwise distances
using modality-specific measures of evolutionary distance.

\paragraph{Proteins (EVE).}
For protein sequences we estimate pairwise distances under the WAG substitution
model~\citep{whelan2001general}, a $20{\times}20$ empirical amino-acid rate matrix
derived from a large corpus of protein families via maximum likelihood.
To account for site-to-site rate variation we use a four-category
discrete-Gamma model (median approximation, shape parameter $\alpha=1.0$)
with category rates $r_1,\dots,r_4$ and equal weights $w_r=\tfrac14$, giving
WAG$+\Gamma$ distances.
Specifically, for two aligned sequences $s_i, s_j$ we maximise over the
evolutionary time (branch length) $t$ the pair-site log-likelihood
\begin{equation}
  \mathcal{L}(t \mid s_i, s_j) = \sum_{k=1}^{L}
    \log \sum_{r=1}^{4} w_r\, P\!\left(s_i^k, s_j^k \mid r\,t\right),
\end{equation}
where $L$ is the number of aligned match columns, $s_i^k$ is the residue of
sequence $s_i$ at column $k$, and $P(x,y \mid \tau)$ is the reversible WAG pair
likelihood for two observed residues at evolutionary time $\tau$, with each
site's probability averaged over the Gamma rate categories.
The resulting distance $d_{ij}=\arg\max_t \mathcal{L}(t \mid s_i, s_j)$ is the
ML estimate of pairwise distance/branch length between sequences $i$ and $j$.

\paragraph{RNA (RfamGen).}
For RNA sequences we operate directly in the CM feature space used by RfamGen~\citep{sumi2024deep}.
Using the CM representation described above, each sequence is encoded by the triplet
$(tr, ss, bp)$.
We define the pairwise distance as the sum of normalised squared $\ell_2$ distances
across the three CM components:
\begin{equation}
  D(s_i, s_j) = \frac{\|tr_i - tr_j\|^2}{2\,|\text{TR}|}
              + \frac{\|ss_i - ss_j\|^2}{2\,|\text{S}|}
              + \frac{\|bp_i - bp_j\|^2}{2\,|\text{P}|},
\end{equation}
where $|\text{TR}|$, $|\text{S}|$, and $|\text{P}|$ are the lengths of the transition,
single-strand, and base-pair components.
For one-hot data, $\|a - b\|^2 = 2$ at every mismatched position, so dividing by
$2 \times \text{length}$ recovers the per-position Hamming rate; thus $D \in [0, 3]$.
This metric measures proximity in the exact space the VAE optimises.

In both cases, pairwise distances are computed \emph{once} at the start of training, and---at each time step---distances from the randomly sampled anchor are used to define the curriculum (Algorithm~\ref{alg:ecl}).

\subsection{Anchor-Based Local-to-Global Curriculum}

At each training step $t$, ECL constructs a mini-batch using an \emph{anchor-and-neighbors}
procedure rather than uniform random sampling.
First, an \textbf{anchor} sequence $a_t$ is drawn from the full training set with
probability proportional to the phylogenetic sequence weights $\pi_s$—the same
reweighting used in the baseline to correct for lineage overrepresentation:
\begin{equation}
  P(a_t = i) \propto \pi_i.
\end{equation}
Then, $B-1$ \textbf{neighbor} sequences are sampled (also proportional to $\pi_s$)
from the $k(t)$-nearest neighbors of $a_t$ in the precomputed distance matrix,
forming a mini-batch of size $B$.

The neighborhood size $k(t)$ expands over training via a schedule that interpolates
log-linearly from an initial size $k_0$ to the full training set $N-1$:
\begin{equation}
  k(t) \;=\; \left\lceil k_0 \cdot \left(\frac{N-1}{k_0}\right)^{s(t)} \right\rceil,
  \quad
  s(t) \;=\; \left(\frac{t-1}{T-1}\right)^{\!\gamma},
  \label{eq:schedule}
\end{equation}
where $T$ is the total number of training steps and $\gamma > 0$ shapes the
expansion rate.
At $t = 1$, $s = 0$ and $k = k_0$; at $t = T$, $s = 1$ and $k = N-1$, so all
sequences are reachable.
Because $k$ grows multiplicatively (log-linearly) rather than additively, the early
training period is spent almost entirely within a tight local neighborhood;
the curriculum then expands rapidly in the latter half of training. Hyperparameter-wise, we use $\gamma=2.0$ throughout. $k(t)$ is also clipped to $[B-1,\, N-1]$ so that the neighborhood
always contains enough candidates to fill a batch without replacement; with
$k_0 \geq B-1$ this clip is inactive except at the endpoints.
Section~\ref{sec:sensitivity} reports the sensitivity of both $k_0$ and $\gamma$.

\begin{algorithm}[t]
  \caption{Evolutionary Curriculum Learning (ECL)}
  \label{alg:ecl}
  \begin{algorithmic}
    \STATE \textbf{Input:} Training sequences $\{s_i\}_{i=1}^N$, pairwise distances $D$,
      weights $\{\pi_i\}$, model $f_\theta$, steps $T$, batch size $B$,
      initial neighborhood $k_0$, exponent $\gamma$
    \STATE \textbf{Output:} Trained parameters $\theta$
    \vspace{2pt}
    \STATE \textit{// Preprocessing (once): sorted neighbor lists}
    \FOR{$i = 1$ {\bfseries to} $N$}
      \STATE $\mathrm{knn}[i] \leftarrow$ indices $\{j : j \neq i\}$ sorted by $D_{ij}$ ascending
    \ENDFOR
    \vspace{2pt}
    \STATE \textit{// Training loop with expanding neighborhood}
    \FOR{$t = 1$ {\bfseries to} $T$}
      \STATE $s \leftarrow \bigl((t-1)/(T-1)\bigr)^{\gamma}$
        \quad \textit{// curriculum progress, $s \in [0,1]$}
      \STATE $k \leftarrow \bigl\lceil k_0\,\bigl((N-1)/k_0\bigr)^{s} \bigr\rceil$
      \STATE $k \leftarrow \mathrm{clip}\bigl(k,\; B-1,\; N-1\bigr)$
        \quad \textit{// enough candidates to fill a batch}
      \STATE Sample anchor $a \sim \mathrm{Categorical}\bigl(\pi / \textstyle\sum_j \pi_j\bigr)$
      \STATE $\mathcal{N} \leftarrow \mathrm{knn}[a]_{1:k}$
        \quad \textit{// $k$ nearest neighbors of $a$}
      \STATE Sample $B-1$ neighbors from $\mathcal{N}$ with probability $\propto \pi$
      \STATE $\mathcal{B} \leftarrow \{a\} \cup \{\text{sampled neighbors}\}$
      \STATE $\theta \leftarrow \theta - \eta\,\nabla_\theta \mathcal{L}_\mathrm{ELBO}(f_\theta;\,\mathcal{B})$
    \ENDFOR
    \STATE \textbf{return} $\theta$
  \end{algorithmic}
\end{algorithm}

\subsection{Relationship to Phylogenetic Reweighting}

Phylogenetic reweighting~\citep{hopf2017mutation} is a common technique for correcting lineage overrepresentation
in MSAs by adjusting the \emph{weight} of
sequences. ECL operates on top of phylogenetic reweighting.
The weights $\pi_s$—which down-weight sequences from overrepresented
lineages~\citep{hopf2017mutation}—govern both anchor selection and neighbor
sampling within $\mathcal{N}$ at every step.

\section{Experiments}
\label{sec:experiments}

We evaluate ECL on both EVE (protein) and RfamGen (RNA) by training baseline and
ECL variants using the configurations described below.
See Appendix~\ref{app:details} for full implementation details and hyperparameters.

\paragraph{Statistical protocol.}
Every configuration is trained with five random seeds ($0$--$4$), and each ECL
variant is compared against the \emph{seed-matched} baseline so that comparisons are
paired.
We report the mean over seeds with a $95\%$ confidence interval, the mean paired
difference with its CI, a $p$-value, and the number of seeds on which the variant
beats its matched baseline.
Every per-task $p$-value reported in this paper is from a paired $t$-test; the only
exception is the pooled 15-run aggregate in Section~\ref{sec:rfamgen}, which uses an
exact binomial sign test.
Two caveats govern how these should be read.
First, the per-task sample size is small ($n=5$ seeds), which makes statistical
significance difficult to attain.
Second, the evaluation sets are small and imbalanced (Table~\ref{tab:clinvar}),
so single-seed differences are noisy; we therefore emphasize consistency of direction
alongside effect size rather than relying on any single point estimate.
Appendix~\ref{app:stats} lists the full set of conditions and the statistical
procedure in detail.

\subsection{Protein Variant Effect Prediction (EVE)}
\label{sec:eve}

\begin{figure}[t]
  \vskip 0.1in
  \begin{center}
    \centerline{\includegraphics[width=\columnwidth]{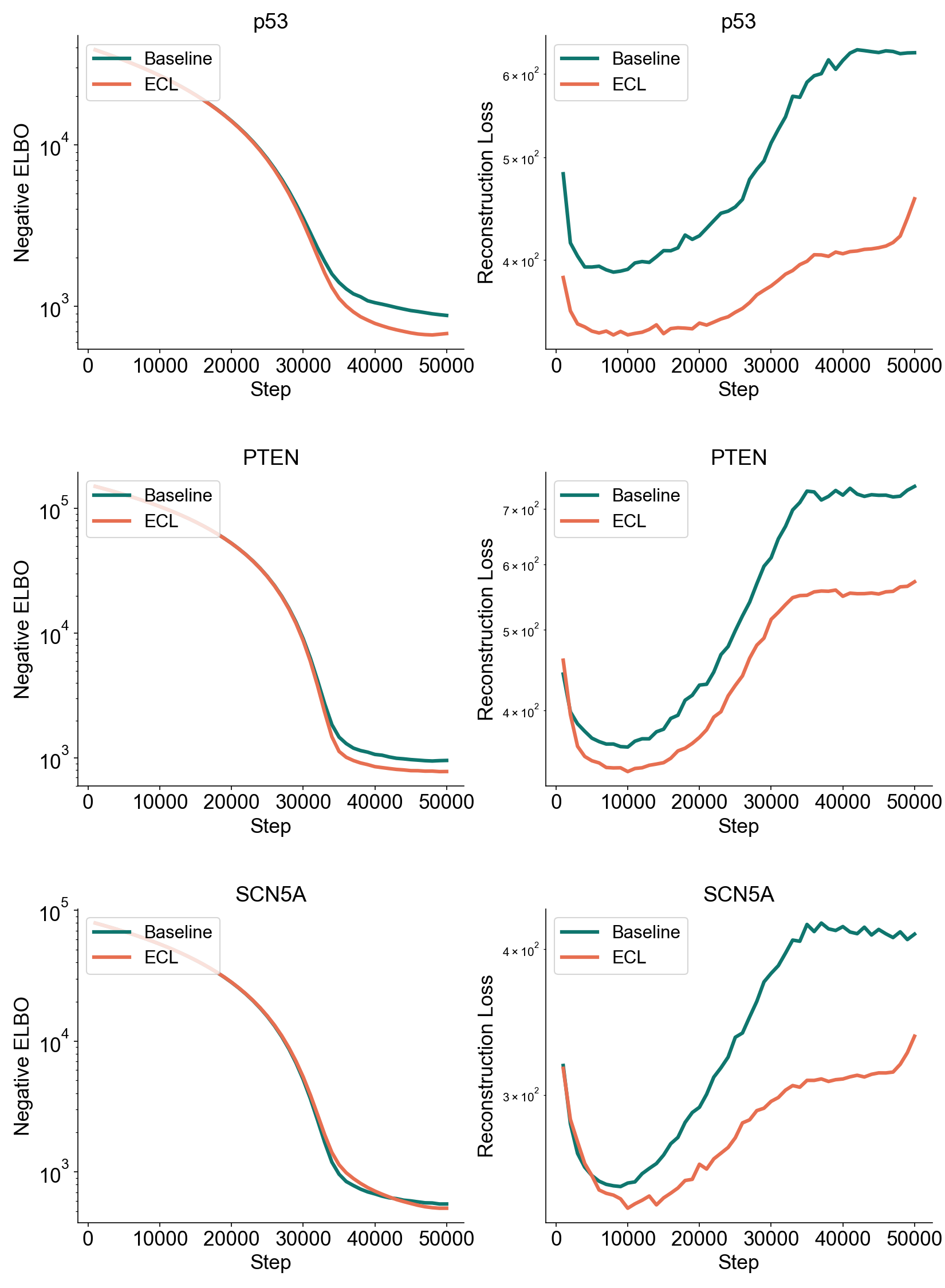}}
    \caption{
      Validation negative ELBO (left column) and reconstruction loss (right column)
      over 50{,}000 training steps for EVE on p53, PTEN, and SCN5A.
      ECL (coral) converges to lower values than the baseline (teal) on all three
      proteins, indicating a better-fitting generative model.
    }
    \label{fig:eve_curves}
  \end{center}
  \vskip -0.1in
\end{figure}

\paragraph{Setup and training dynamics.}
We train EVE on three disease-relevant proteins: p53 ($n=3,668$ MSA sequences), PTEN
($n=1,197$), and SCN5A ($n=11,848$), with five seeds per configuration.
Figure~\ref{fig:eve_curves} shows validation negative ELBO and reconstruction loss
over training for a representative seed.
ECL converges to lower values for both metrics, with particularly large gaps for p53
and PTEN.

\paragraph{Variant pathogenicity classification.}
Table~\ref{tab:clinvar} reports AUROC for classifying ClinVar pathogenic versus benign
variants using the EVE score (the posterior probability of assignment to the pathogenic cluster of a two-component GMM fit to the per-variant evolutionary indices), averaged over five seeds.
On p53, ECL raises mean AUROC from $0.9814$ to $0.9889$ ($+0.0075$, 95\% CI
$[-0.0012, +0.0161]$, $t$-test $p=0.075$), winning on 4 of 5 seeds.
On PTEN, ECL attains $1.0000$ on \emph{every} seed, while the baseline averages
$0.9054$ with a very wide interval: four baseline seeds score $\geq 0.99$ but one
collapses to $0.5405$.
The improvement on PTEN is therefore best described as a gain in \emph{stability}
rather than a large shift in central tendency, and the paired difference is not
significant ($p=0.36$) precisely because the baseline variance is so large.
Figure~\ref{fig:protein_auroc} shows all conditions with per-seed values.
Figure~\ref{fig:variant_separation} visualizes the score distributions, where
ECL-trained models separate pathogenic from benign variants more cleanly.

\paragraph{Label imbalance and a position-only baseline.}
The ClinVar sets available for these MSAs are small and heavily imbalanced
(Table~\ref{tab:clinvar}); PTEN in particular retains only \emph{two} benign variants
after intersecting ClinVar labels with the positions the MSA actually models.
An AUROC of $1.000$ on PTEN therefore means only that both benign variants rank below
all $111$ pathogenic ones.
For SCN5A the intersection leaves $23$ labeled variants, \emph{all} pathogenic, so
AUROC is undefined; we consequently report SCN5A only through training curves
(Figure~\ref{fig:eve_curves}) and exclude it from Table~\ref{tab:clinvar}.
To test the concern that pathogenic variants simply cluster in a few regions—so that
a trivial positional classifier could match the model—we fit a cross-validated
position-only classifier ($k$-NN on residue index).
It reaches AUROC $0.841$ on p53 and $0.381$ on PTEN (means over the five seeds),
both far below the VAE scores, confirming that the models capture substitution-level
signal rather than a spatial artifact.

\begin{table}[t]
  \caption{ClinVar variant classification AUROC, mean over five seeds.
    $\Delta$ is the mean seed-paired difference vs.\ the baseline.
    ``Position-only'' is a cross-validated $k$-NN classifier on residue index.
    SCN5A is omitted: after intersecting with the modeled MSA positions its labeled
    set contains 23 variants, all pathogenic, leaving AUROC undefined.}
  \label{tab:clinvar}
  \vskip 0.05in
  \begin{center}
    \begin{small}
        \begin{tabular}{llccc}
          \toprule
          Protein & Method & AUROC & $\Delta$ & $p$ \\
          \midrule
          p53            & Baseline      & 0.9814 & --- & --- \\
          (130p / 27b)   & ECL           & \textbf{0.9889} & $+0.0075$ & 0.075 \\
                         & Fixed-$k$     & 0.9483 & $-0.0332$ & 0.0001 \\
                         & Position-only & 0.8414 & --- & --- \\
          \midrule
          PTEN           & Baseline      & 0.9054 & --- & --- \\
          (111p / 2b)    & ECL           & \textbf{1.0000} & $+0.0946$ & 0.36 \\
                         & Fixed-$k$     & 1.0000 & $+0.0946$ & 0.36 \\
                         & Position-only & 0.3811 & --- & --- \\
          \bottomrule
        \end{tabular}
    \end{small}
  \end{center}
  \vskip -0.1in
\end{table}

\subsection{RNA Family Sequence Generation (RfamGen)}
\label{sec:rfamgen}

\paragraph{Setup.}
We train RfamGen on three Rfam~\citep{kalvari2020rfam} families spanning diverse RNA
classes: RF00169 ($n=6,168$), RF00234 ($n=900$), and RF00638 ($n=2,999$).
We use the original RfamGen architecture~\citep{sumi2024deep} and training
hyperparameters, with two changes for ECL: the curriculum sampling procedure in
Algorithm~\ref{alg:ecl}, and a non-cyclic annealing schedule for the $\beta$ weight
on the KL divergence term to avoid repeatedly resetting $\beta$ to zero.
Because these two changes are confounded if applied together, we ablate them
separately in Section~\ref{sec:ablation}.
After training, we sample 1{,}000 sequences per model and score them with the
ground-truth CM for that RNA family to compute bit scores, which measure how well a
generated sequence aligns with the family and assign higher values to better matches
\cite{sumi2024deep}.

\begin{figure}[t]
  \vskip 0.1in
  \begin{center}
    \centerline{\includegraphics[width=0.92\columnwidth]{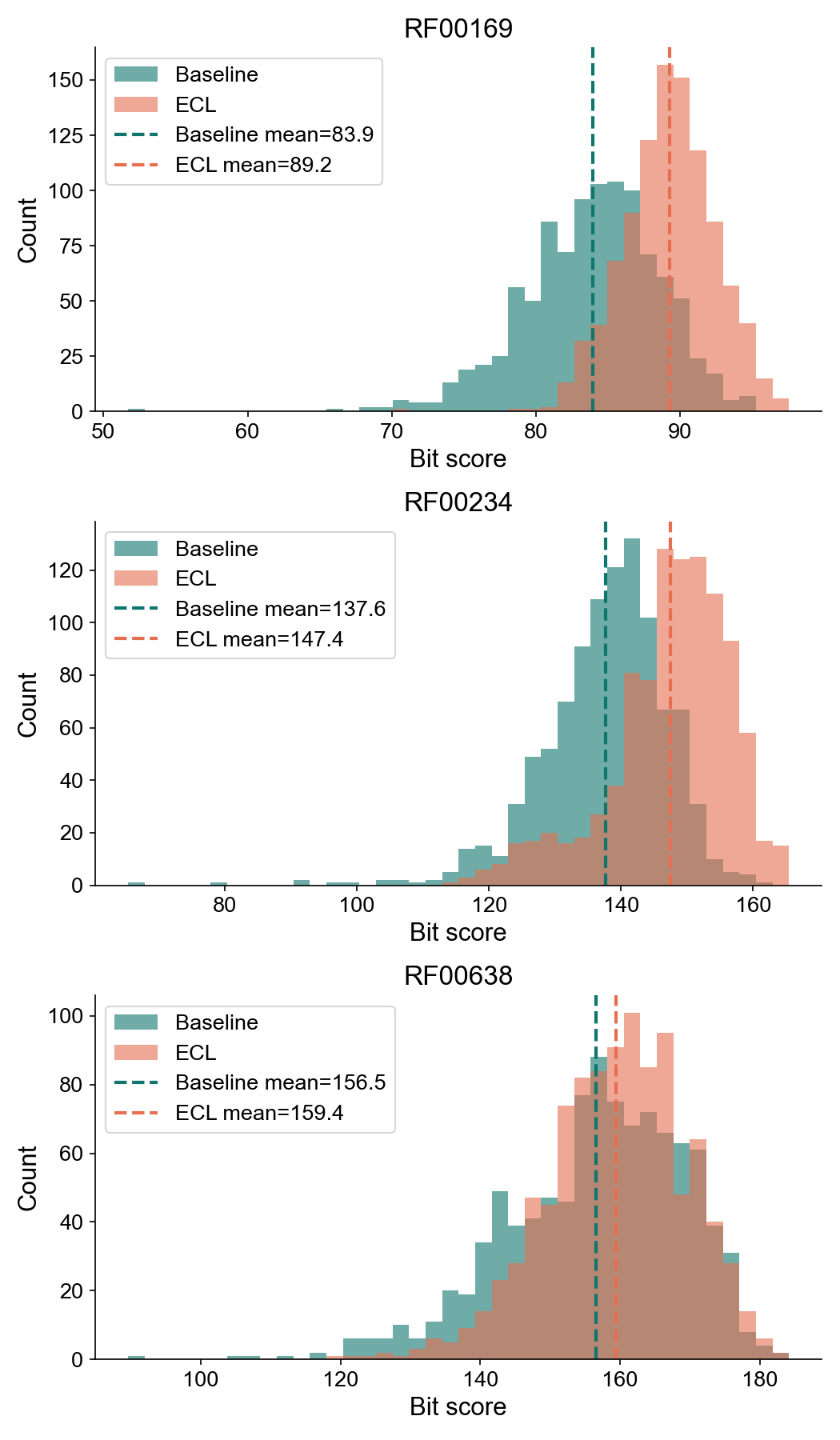}}
    \caption{
      CM bit score distributions for 1{,}000 sequences sampled from a single
      representative seed of the baseline (teal) and ECL (coral) RfamGen models on
      each of three RNA families. Distributions therefore reflect one model per
      condition, not an average over the five seeds; per-seed and seed-averaged
      results are reported in Figure~\ref{fig:rna_bitscore} and
      Table~\ref{tab:rna_full}.
      Dashed lines mark mean bit scores.
      ECL consistently produces higher-scoring sequences across all three families.
    }
    \label{fig:rfamgen}
  \end{center}
  \vskip -0.1in
\end{figure}

\paragraph{Results.}
Figure~\ref{fig:rfamgen} shows CM bit score distributions for a single
representative seed per condition, and
Figure~\ref{fig:rna_bitscore} reports every condition across five seeds, with
checkpoints selected by validation VAE loss.
Averaged over seeds, ECL yields higher mean bit scores on all three families:
$86.0$ vs.\ $83.0$ for RF00169 ($+3.7\%$, improved on 5/5 seeds, $p=0.077$),
$150.2$ vs.\ $139.6$ for RF00234 ($+7.5\%$, 3/5, $p=0.33$), and
$153.7$ vs.\ $152.3$ for RF00638 ($+0.9\%$, 4/5, $p=0.77$).
Individually these per-family effects are underpowered at $n=5$.
Aggregating the 15 seed-matched comparisons, ECL wins 12 of 15 runs, which an exact
sign test rejects at $p = 0.035$.
We report this statistic with an explicit caveat about its scope.
The unit
of inference for this test is the training \emph{run}, not the RNA family: it supports
the claim that ECL reliably improves training runs on these three families, not that
it improves RNA families in general.
This is a limitation of the design: with only three families, a family-level test cannot establish significance even if every family improves (a two-sided sign test over three families floors at $p = 0.25$).
Establishing family-level generality would require substantially more RNA families and compute resources.
Appendix~\ref{app:rna_full} reports every condition and family.

\subsection{Isolating the Contribution of the Curriculum}
\label{sec:ablation}

ECL changes two things relative to uniform sampling: batches become \emph{local}
(drawn from an anchor's neighborhood at all), and that locality \emph{expands} over
training. The baseline-versus-ECL comparisons above establish that this combination improves on uniform sampling. Now, we isolate the effect of the locality expansion with a fixed-$k$ variant that retains anchor-based local sampling but holds $k(t) = k_0$ for the entire run.

On p53, fixed-$k$ is not merely neutral but clearly harmful: mean AUROC falls to
$0.9483$, $-0.0332$ below the baseline (95\% CI $[-0.0382, -0.0281]$, $p = 0.0001$),
losing on all five seeds. Training only on tight neighborhoods thus starves the model
of the global variation it needs, and the benefit of ECL depends on the expansion
rather than on locality alone. On PTEN, fixed-$k$ also reaches $1.000$ AUROC, but this is uninformative: with only two benign variants, many conditions (including ECL and most baseline seeds) score $1.000$, so the metric is saturated and cannot distinguish them. The evidence that expansion matters therefore comes from p53. Across RNA families fixed-$k$ improves the pooled bit score by only
$2.3\%$ (7/15 runs) compared to $5.0\%$ (12/15) for full ECL.
Figure~\ref{fig:ablation} summarizes both modalities.

\paragraph{Decoupling the annealing schedule.}
For RfamGen, ECL is naturally paired with a non-cyclic KL schedule, so we ran all
four combinations of \{baseline, ECL\} sampling $\times$ \{cyclic, non-cyclic\}
annealing to determine which component drives the result.
Changing only the annealing (baseline $+$ non-cyclic) does \emph{not} reproduce ECL:
pooled across families it \emph{lowers} the mean bit score by $5.0\%$, and it is
substantially harmful on RF00234 ($-11.5\%$) and RF00638 ($-10.6\%$) while helping
only RF00169 ($+3.4\%$).
Conversely, ECL combined with the original cyclic schedule still improves over the
baseline ($+5.7\%$ pooled).
The curriculum sampler, not the annealing schedule, is therefore responsible for the
improvement. ECL helps under either annealing schedule, but the non-cyclic schedule
is the more consistent of the two (12/15 vs.\ 8/15 winning runs for the cyclic schedule).

\subsection{Sensitivity to $k_0$ and $\gamma$}
\label{sec:sensitivity}

Figure~\ref{fig:sensitivity} sweeps the initial neighborhood $k_0$ and the schedule
exponent $\gamma$ around their defaults.
On p53 every setting stays at or above the baseline
($k_0 \in \{255, 512, 1024\} \rightarrow 0.9889, 0.9846, 0.9892$;
$\gamma \in \{1, 2, 4\} \rightarrow 0.9893, 0.9889, 0.9830$), and across RNA families
the pooled improvement varies only between $+4.7\%$ and $+6.1\%$ over the same grid.
ECL is therefore not delicately tuned.

The one systematic failure is informative. On PTEN, $k_0 = 1024$ drops mean AUROC to
$0.7541$ and wins on only 1 of 5 seeds. PTEN's training set has $N-1 = 1076$, so
$k_0 = 1024$ makes the initial neighborhood nearly the entire alignment: the
curriculum is effectively removed and sampling is global from the first step.
The same $k_0$ is harmless on p53, whose alignment is roughly three times larger and
for which $1024$ is still a genuinely local neighborhood.

\subsection{Computational Cost and Scalability}
\label{sec:cost}

ECL adds no per-step training overhead: sampling an anchor and its neighbors is
$O(B)$ per step, exactly as for weighted uniform sampling.
The only additional cost is the pairwise distance matrix, computed once and cached,
then reused across every run, seed, and hyperparameter setting in this paper.
This preprocessing is $O(N^2)$ in the number of sequences for both protein and RNA modalities, since all pairs are compared.

Because this term is quadratic, it is the main obstacle to applying ECL to much
deeper alignments.
We note that the curriculum consumes only \emph{ordered} neighbor lists rather than
exact distances, so any monotone proxy would suffice; identifying cheaper such
proxies is left to future work.

\subsection{Positioning Relative to Other Predictors}
\label{sec:positioning}

ECL is a training-strategy intervention rather than a new model class, so the comparison reported above is baseline-EVE versus ECL-EVE with identical architecture,
data splits, optimizer, and compute budget. We therefore do not claim state-of-the-art variant effect prediction.
ECL is instead orthogonal to, and composable with, any MSA-based generative
predictor: it changes only the order in which training sequences are visited.

\section{Discussion and Conclusion}
\label{sec:conclusion}

We have introduced Evolutionary Curriculum Learning, a simple and effective training
strategy for MSA-based biological sequence VAEs.
By sampling training sequences by evolutionary distance from sampled anchors and
expanding the neighborhood according to a power-law schedule, ECL provides
evolutionary structure as an inductive bias at no architectural cost.

Across two biological modalities—protein and RNA—and two architecturally distinct
models, ECL improves downstream task performance in a consistent direction: variant
classification AUROC for disease prediction and CM bit scores for RNA generation
quality.

Several limitations should be recognized, however.
Effect sizes are modest relative to seed variance: with five seeds per configuration,
most individual comparisons do not reach $p<0.05$, and our strongest evidence is the
consistency of the direction together with the ablations rather than any single
headline number.
The aggregate RNA statistic (12 of 15 runs, $p=0.035$) treats the training run as the
unit of inference; family-level tests over
three families cannot achieve significance.
The clinical evaluation sets are also small and imbalanced---PTEN
retains two benign variants and SCN5A none once labels are intersected with the
modeled MSA positions---so protein AUROC is a coarse instrument here and saturates
easily. Larger and better-balanced benchmarks are needed to measure the effect
precisely.
The evaluation covers three proteins and three RNA families, which is too few to
support strong claims of generality.
Finally, the curriculum requires an $O(N^2)$ preprocessing step
(Section~\ref{sec:cost}).
Future work should evaluate ECL at larger benchmark scale, explore alternative
evolutionary distance metrics, and test whether the same local-to-global principle
helps non-VAE sequence models.


\section*{Impact Statement}

This paper advances training methods for biological sequence generative models,
with potential applications in clinical variant interpretation and RNA engineering
for therapeutics. Results should be interpreted
alongside experimental and clinical evidence.

\section*{Code Availability}

The code for this project is publicly available at:
\url{https://github.com/KentoNishi/icml26-ecl}.
\bibliography{references}
\bibliographystyle{icml2026}

\newpage
\appendix
\onecolumn

\section{Implementation Details}
\label{app:details}

\subsection{EVE}
EVE models use a Bayesian VAE with encoder MLP [512,128], latent dimension 16, and
Bayesian decoder MLP [128,512], with no convolution, temperature scaler, sparsity, or dropout.
MSAs for p53, PTEN, and SCN5A were retrieved from UniRef following the original EVE
pipeline; each MSA was split 90/10 into train/validation sets.
All models were trained for 50{,}000 steps with batch size $B = 256$ using Adam with
learning rate $3{\times}10^{-4}$ and no learning-rate scheduler.
WAG$+\Gamma$ pairwise distances (shape $\alpha = 1.0$, 4 rate categories) were
precomputed once before training.
For ECL we set $k_0 = 255$ (approximately $1{\times}B$), $\gamma = 2.0$.
Phylogenetic sequence reweighting (threshold $\theta_\text{seq} = 0.2$, as in the
original EVE paper) was applied to both anchor selection and neighbor sampling at
every step.
Evolutionary indices were computed with 200 Monte Carlo samples; ClinVar AUROC was
computed from the pathogenic-component posterior of a two-component GMM fit to the computed evolutionary indices.

\subsection{RfamGen}
CM-feature pairwise distances were precomputed from the training HDF5 files as
described in Section~\ref{sec:method}.
For ECL we set $k_0 = 240 = 30{\times}B$ and $\gamma = 2.0$.
Sequence reweighting (threshold $\theta_\text{seq} = 0.1$, per the original RfamGen
protocol) was applied in the loss; ECL also used these weights for anchor and neighbor sampling. Data was split into 70/15/15 train/val/test sets for each RNA family.
RfamGen models follow the architecture of \citet{sumi2024deep}: a 16-dimensional
latent space with a convolutional CM-structured encoder/decoder.
We trained on RF00169, RF00234, and RF00638 from Rfam 14.7~\citep{kalvari2020rfam}
with batch size $B = 8$ for up to 200 epochs with early stopping (training stopped once validation loss increases for 3 successive epochs). The checkpoint chosen for bit score evaluation was the best validation loss epoch checkpoint up until the stopped epoch.
Bit scores were computed for 1{,}000 sequences sampled from the best-validation-epoch
checkpoint of each model by aligning generated sequences to the ground-truth Rfam CM
using Infernal~\citep{nawrocki2013infernal}.

\subsection{Hyperparameters}

\begin{table}[h]
  \caption{ECL hyperparameters used in all experiments.}
  \label{tab:hparams}
  \vskip 0.05in
  \begin{center}
    \begin{small}
      \begin{sc}
        \begin{tabular}{lcc}
          \toprule
          Parameter & EVE (protein) & RfamGen (RNA) \\
          \midrule
          Batch size $B$      & 256  & 8   \\
          Initial $k_0$       & 255  & 240 \\
          $k_0 / B$           & 1.0  & 30  \\
          Exponent $\gamma$   & 2.0  & 2.0 \\
          Training steps $T$  & 50{,}000 & $\leq$200 epochs \\
          Distance metric     & WAG$+\Gamma$ & CM-feature $D_H$ \\
          \bottomrule
        \end{tabular}
      \end{sc}
    \end{small}
  \end{center}
\end{table}

The asymmetry in $k_0/B$ between EVE and RfamGen (1 vs.\ 30) reflects the difference
in training batch sizes.

\subsection{Seeds, Conditions, and Statistical Procedure}
\label{app:stats}

Every condition below was trained with five seeds ($0$--$4$) for both modalities.
For EVE the conditions are: baseline; ECL ($k_0{=}255$, $\gamma{=}2$); fixed-$k$
($k_0{=}255$, no expansion); and the sweep $k_0 \in \{512, 1024\}$ at $\gamma{=}2$
and $\gamma \in \{1, 4\}$ at $k_0{=}255$.
For RfamGen the conditions are: baseline (cyclic annealing); ECL ($k_0{=}240$,
$\gamma{=}2$, non-cyclic); fixed-$k$; ECL with cyclic annealing; baseline with
non-cyclic annealing; and the sweep $k_0 \in \{120, 480\}$ at $\gamma{=}2$ and
$\gamma \in \{1, 4\}$ at $k_0{=}240$.
This is $7 \times 5 = 35$ EVE runs per protein and $9 \times 5 = 45$ RfamGen runs per
family.

Each ECL variant is compared against the baseline trained with the \emph{same} seed,
making all comparisons paired. We report the mean over seeds, the mean paired difference, and a paired $t$-test.
The paired $t$-test assumes approximately normal paired differences, an assumption
that cannot be verified at $n=5$ and is sensitive to outliers (e.g.\ the collapsed
PTEN baseline seed); we therefore treat the per-task $t$-test $p$-values as
descriptive rather than confirmatory, and note that at $n=5$ it is difficult to attain statistical significance---the exact two-sided Wilcoxon signed-rank and sign
tests both floor at $2/2^5=0.0625$, for example.
We reserve exact binomial sign tests for aggregating runs across RNA families, where
between-family differences in bit-score scale make a pooled $t$-test on absolute
values inappropriate; the sign statistic is scale-free, and each seed-matched pair
forms its own exchangeability block, so the aggregate test is a valid stratified
permutation test of the global null.
Its unit of inference, however, is the training run rather than the family.
For both the protein and RNA evaluations, the confidence intervals reported in
Tables~\ref{tab:clinvar} and~\ref{tab:rna_full} are across-seed $95\%$ intervals:
we average each run's point estimate over the five seeds and form a Student-$t$
interval from the seed-to-seed variability.

\subsection{Full RNA Results}
\label{app:rna_full}

Table~\ref{tab:rna_full} lists every RfamGen condition. Condition names follow
Section~\ref{sec:ablation}: \texttt{ECL cyclic} is the ECL sampler with the original
cyclic annealing, and \texttt{Base non-cyclic} is the unmodified sampler with the
non-cyclic schedule.

\begin{table}[h]
  \caption{CM bit score for every RfamGen condition, mean $\pm$ 95\% CI over five
    seeds. $\Delta$ and \% are relative to the seed-matched baseline; ``wins''
    counts seeds on which the condition beats its matched baseline.}
  \label{tab:rna_full}
  \vskip 0.05in
  \begin{center}
    \begin{small}
      \begin{tabular}{llccccc}
        \toprule
        Family & Condition & Bit score & $\Delta$ & \% & $p$ & wins \\
        \midrule
        RF00169 & Baseline        & 83.0 [79.8, 86.2]    & --     & --      & --    & -- \\
                & ECL             & 86.0 [83.6, 88.4]    & $+3.1$ & $+3.7$  & 0.077 & 5/5 \\
                & Fixed-$k$       & 82.6 [77.9, 87.4]    & $-0.3$ & $-0.4$  & 0.849 & 2/5 \\
                & ECL cyclic      & 86.5 [85.3, 87.6]    & $+3.5$ & $+4.2$  & 0.047 & 4/5 \\
                & Base non-cyclic & 85.7 [80.6, 90.9]    & $+2.8$ & $+3.4$  & 0.371 & 4/5 \\
                & $k_0{=}120$     & 86.0 [82.3, 89.6]    & $+3.0$ & $+3.6$  & 0.165 & 4/5 \\
                & $k_0{=}480$     & 85.4 [82.5, 88.4]    & $+2.5$ & $+3.0$  & 0.099 & 5/5 \\
                & $\gamma{=}1$    & 86.1 [81.5, 90.6]    & $+3.1$ & $+3.7$  & 0.285 & 4/5 \\
                & $\gamma{=}4$    & 85.6 [82.2, 89.0]    & $+2.7$ & $+3.2$  & 0.107 & 4/5 \\
        \midrule
        RF00234 & Baseline        & 139.6 [112.5, 166.7] & --      & --       & --    & -- \\
                & ECL             & 150.2 [144.4, 155.9] & $+10.5$ & $+7.5$   & 0.333 & 3/5 \\
                & Fixed-$k$       & 147.1 [141.9, 152.2] & $+7.4$  & $+5.3$   & 0.535 & 2/5 \\
                & ECL cyclic      & 148.0 [142.1, 153.9] & $+8.4$  & $+6.0$   & 0.412 & 2/5 \\
                & Base non-cyclic & 123.5 [109.2, 137.8] & $-16.1$ & $-11.5$  & 0.276 & 1/5 \\
                & $k_0{=}120$     & 153.2 [145.6, 160.8] & $+13.6$ & $+9.7$   & 0.208 & 4/5 \\
                & $k_0{=}480$     & 145.5 [136.9, 154.2] & $+5.9$  & $+4.2$   & 0.536 & 2/5 \\
                & $\gamma{=}1$    & 145.5 [136.0, 154.9] & $+5.8$  & $+4.2$   & 0.674 & 2/5 \\
                & $\gamma{=}4$    & 151.5 [147.0, 155.9] & $+11.8$ & $+8.5$   & 0.302 & 3/5 \\
        \midrule
        RF00638 & Baseline        & 152.3 [137.7, 166.9] & --      & --       & --    & -- \\
                & ECL             & 153.7 [141.9, 165.4] & $+1.3$  & $+0.9$   & 0.773 & 4/5 \\
                & Fixed-$k$       & 150.7 [131.7, 169.7] & $-1.6$  & $-1.1$   & 0.819 & 3/5 \\
                & ECL cyclic      & 157.8 [149.0, 166.6] & $+5.5$  & $+3.6$   & 0.529 & 2/5 \\
                & Base non-cyclic & 136.2 [123.0, 149.4] & $-16.1$ & $-10.6$  & 0.143 & 1/5 \\
                & $k_0{=}120$     & 150.6 [137.9, 163.3] & $-1.7$  & $-1.1$   & 0.853 & 2/5 \\
                & $k_0{=}480$     & 158.4 [148.3, 168.5] & $+6.1$  & $+4.0$   & 0.339 & 4/5 \\
                & $\gamma{=}1$    & 161.6 [156.1, 167.1] & $+9.3$  & $+6.1$   & 0.266 & 4/5 \\
                & $\gamma{=}4$    & 150.7 [132.4, 169.1] & $-1.6$  & $-1.0$   & 0.873 & 3/5 \\
        \bottomrule
      \end{tabular}
    \end{small}
  \end{center}
\end{table}

Pooling the 15 seed-matched comparisons per condition across the three families gives
mean relative improvements of $+5.0\%$ for ECL (12/15 winning runs, sign test
$p = 0.035$), $+5.7\%$ for ECL with cyclic annealing (8/15), $+2.3\%$ for fixed-$k$
(7/15), and $-5.0\%$ for the non-cyclic annealing-only variant (6/15).

\section{Per-Condition Results Across Seeds}
\label{app:figures}

\begin{figure}[h]
  \begin{center}
    \includegraphics[width=\textwidth]{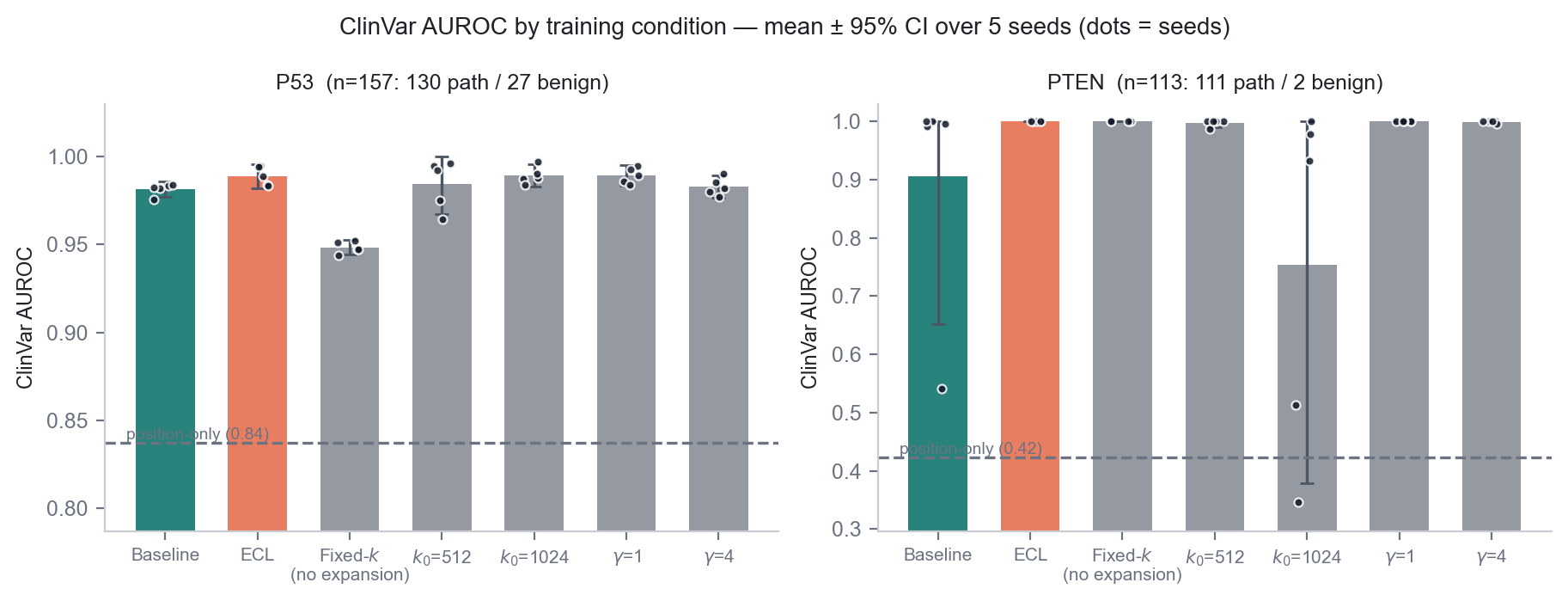}
    \caption{
      ClinVar AUROC by training condition for p53 and PTEN.
      ECL is stable at $1.000$ on PTEN whereas one baseline seed collapses to $0.54$.
      Removing the expansion (fixed-$k$) is clearly harmful on p53, and setting
      $k_0 \to N-1$ on PTEN ($k_0{=}1024$, where $N-1{=}1076$) destroys the curriculum
      and destabilizes the model.
      Dashed line: cross-validated position-only classifier.
    }
    \label{fig:protein_auroc}
  \end{center}
\end{figure}

\begin{figure}[h]
  \begin{center}
    \includegraphics[width=\textwidth]{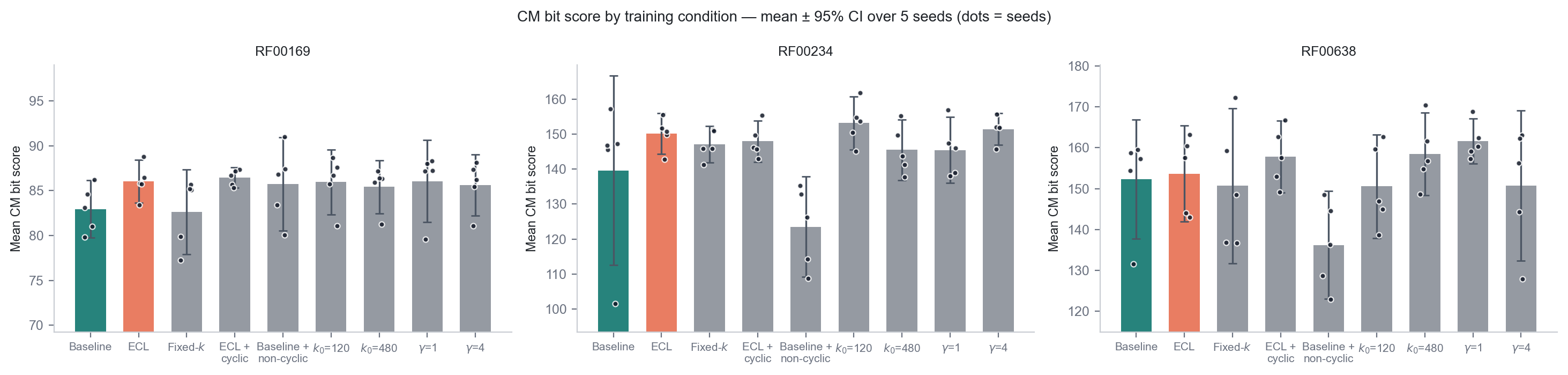}
    \caption{
      Mean CM bit score by training condition for each RNA family.
      ECL improves over the baseline on all three families, whereas the
      annealing-only variant (baseline $+$ non-cyclic) is markedly harmful on
      RF00234 and RF00638.
    }
    \label{fig:rna_bitscore}
  \end{center}
\end{figure}

\begin{figure}[h]
  \begin{center}
    \includegraphics[width=\textwidth]{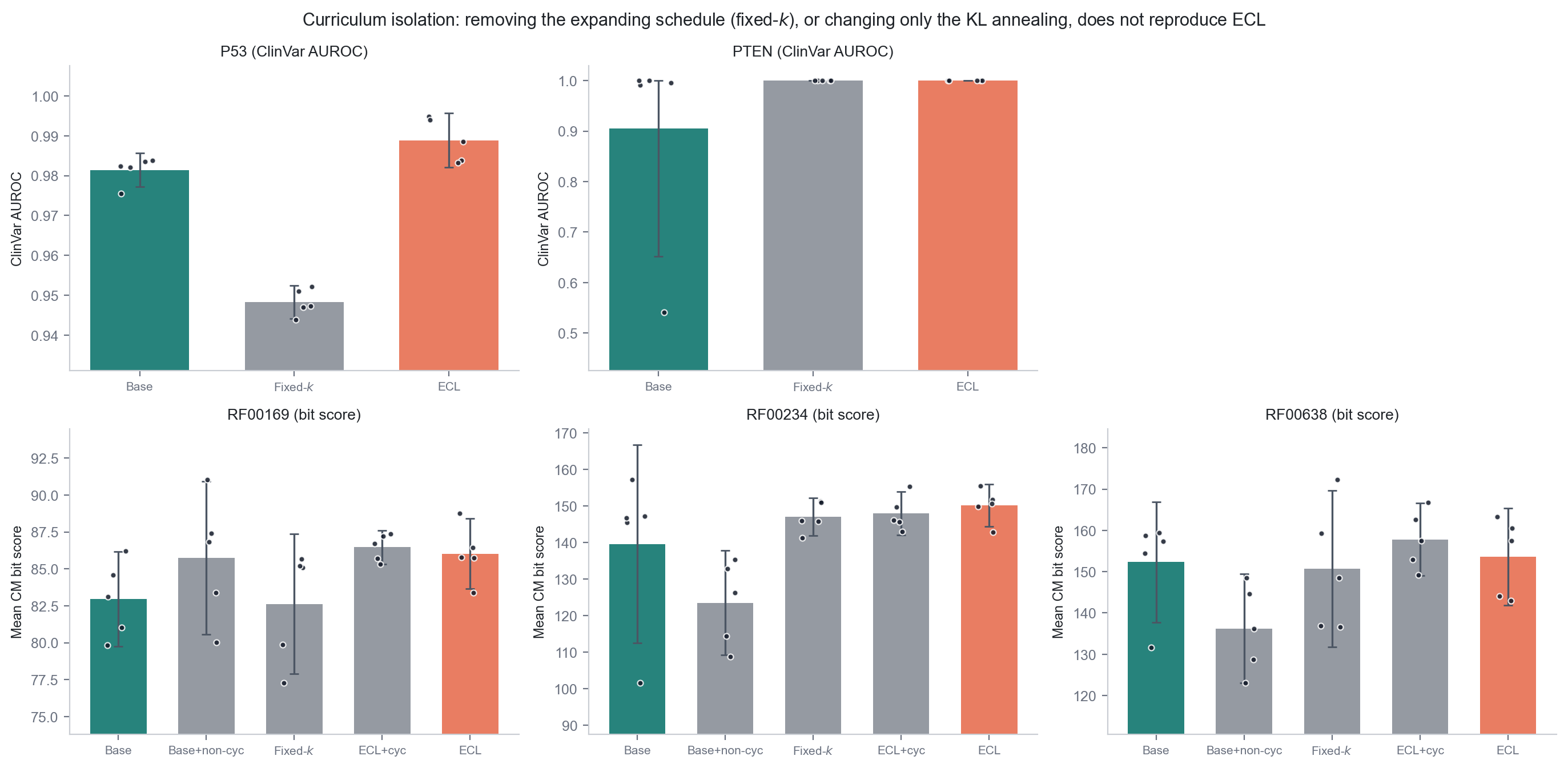}
    \caption{
      Curriculum isolation across both modalities.
      Removing the expansion (fixed-$k$) forfeits the gain and is significantly worse
      than the baseline on p53, and changing only the KL annealing
      (baseline $+$ non-cyclic) is harmful on two of three RNA families.
      On PTEN multiple conditions saturate at $1.000$ because only two benign variants
      remain in the evaluation set, so that panel does not discriminate well between different model conditions.
    }
    \label{fig:ablation}
  \end{center}
\end{figure}

\begin{figure}[h]
  \begin{center}
    \includegraphics[width=\textwidth]{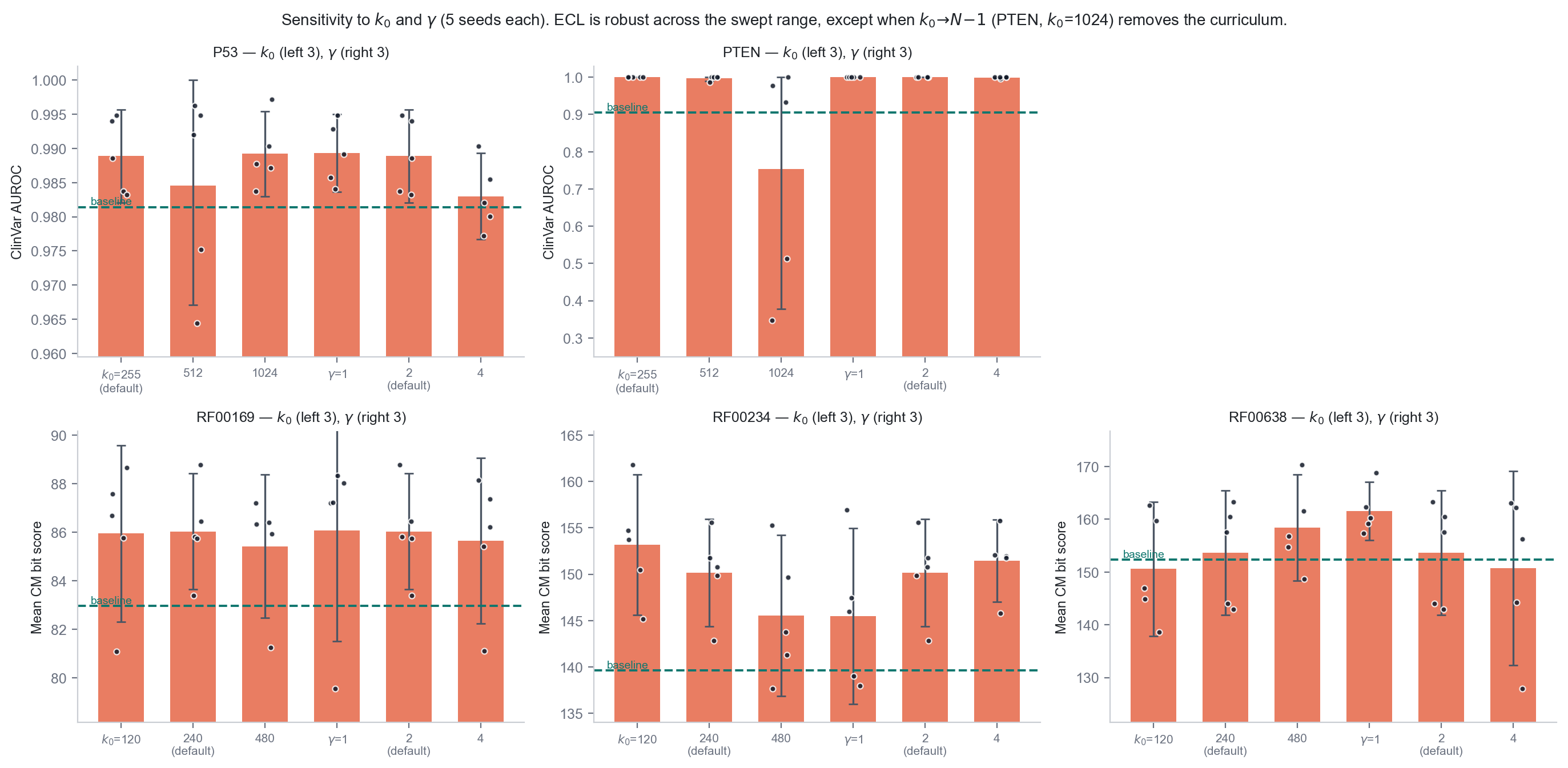}
    \caption{
      Sensitivity to $k_0$ and $\gamma$.
      Performance is stable across the swept range; the exception is PTEN with
      $k_0 = 1024 \approx N-1$, where the neighborhood spans essentially the whole
      alignment and the curriculum no longer exists.
      Dashed line: the seed-matched baseline mean.
    }
    \label{fig:sensitivity}
  \end{center}
\end{figure}

This section shows Figures \ref{fig:protein_auroc}-\ref{fig:sensitivity}, referenced from
Sections~\ref{sec:eve}--\ref{sec:sensitivity}.
Throughout, bars give the mean over five seeds with a $95\%$ confidence interval,
overlaid dots are the individual seeds, teal marks the baseline and coral the ECL
configuration used in the main results; grey marks ablation and sweep conditions.

\section{EVE score distributions for p53 and PTEN}
\label{app:eve_additional}

\begin{figure}[t]
  \vskip 0.1in
  \begin{center}
    \centerline{\includegraphics[width=0.9\columnwidth]{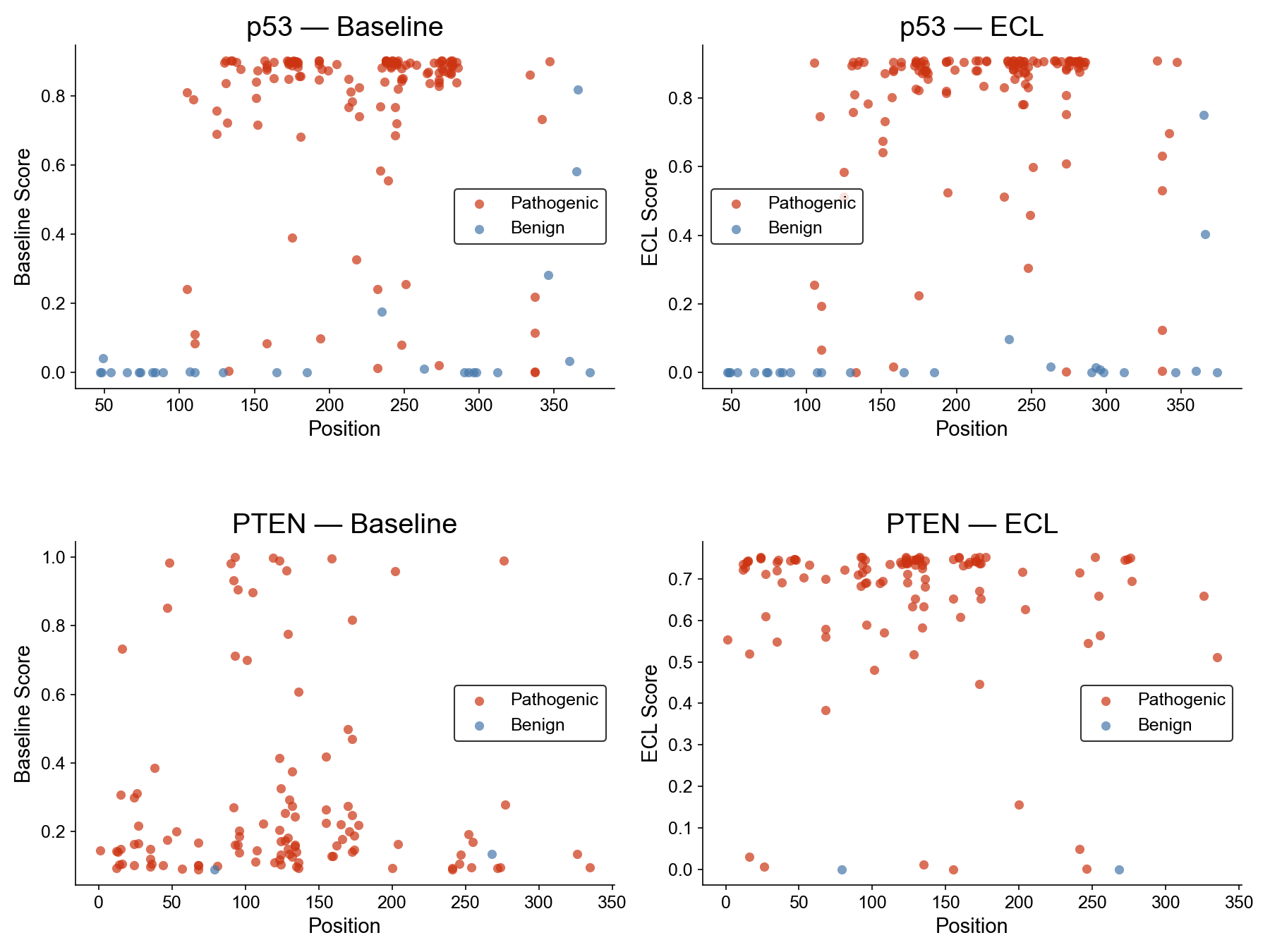}}
    \caption{
      Distribution of EVE scores for variants of two proteins: p53 (top) and PTEN (bottom). Models on the left were trained using the standard baseline paradigm, while those on the right were trained using ECL. Greater separation of benign and pathogenic variants is observed for ECL-trained EVE scores.
    }
    \label{fig:variant_separation}
  \end{center}
  \vskip -0.1in
\end{figure}

See Figure \ref{fig:variant_separation}.

\section{RfamGen Optimization Dynamics for Bit score}
\label{app:rfamgen_curves}

\begin{figure}[t]
  \vskip 0.1in
  \begin{center}
    \centerline{\includegraphics[width=0.6\columnwidth]{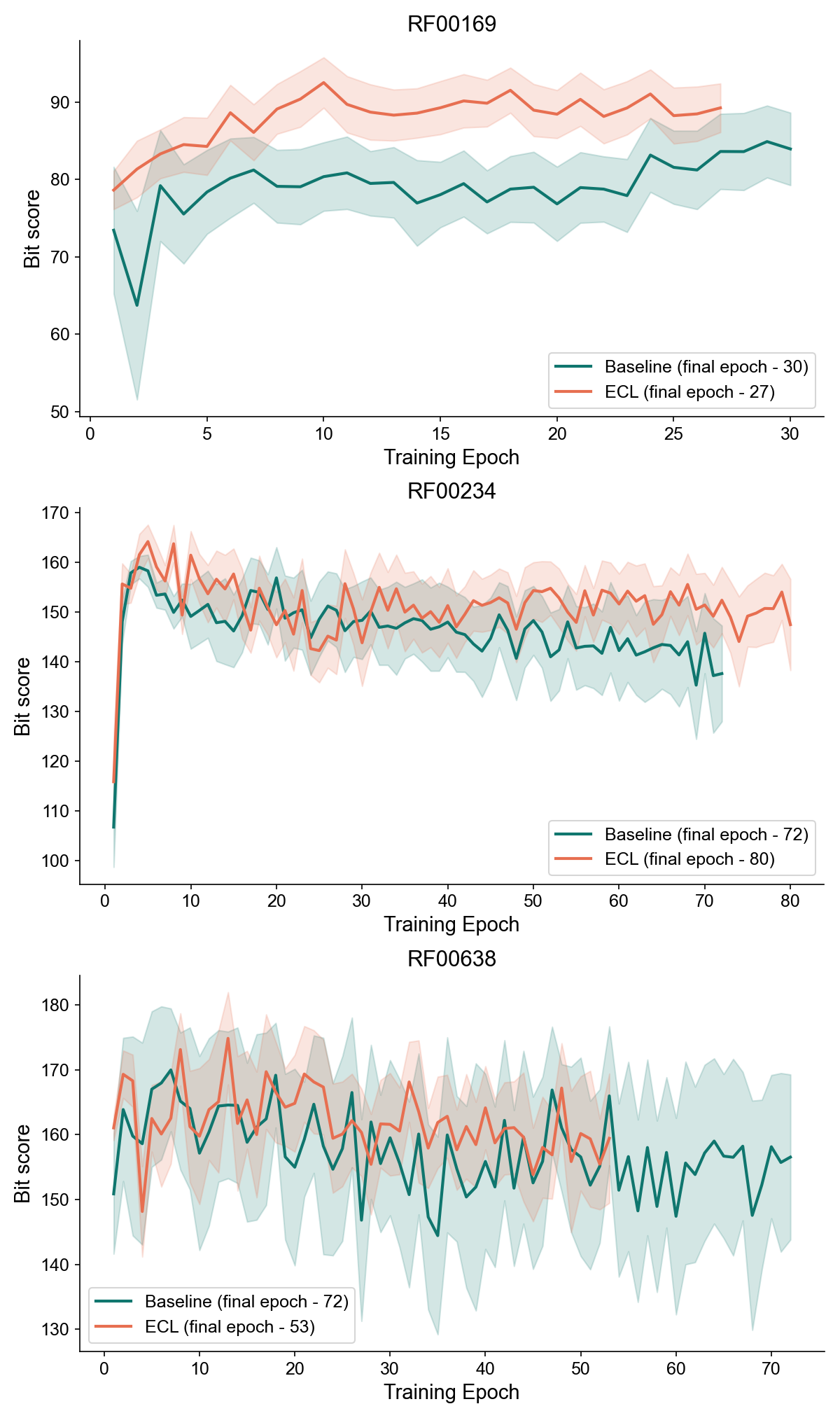}}
    \caption{
          Optimization dynamics of the CM bit score metric over training epochs.
          This post-hoc analysis was performed after checkpoint selection for the experiments in Figure~\ref{fig:rfamgen}, which avoids data leakage. The plot shows the mean and
          standard deviation of bit scores for 1,000 sampled sequences from baseline and
          ECL-trained RfamGen models at each epoch up to the checkpoint selected for
          evaluation in Figure~\ref{fig:rfamgen}.
        }
    \label{fig:rfamgen_epochs}
  \end{center}
  \vskip -0.1in
\end{figure}

We were also interested in the optimization dynamics of the bit score metric across
training epochs. This post-hoc analysis was performed only after checkpoint selection
for Figure~\ref{fig:rfamgen}, which was based on validation loss, to avoid data
leakage. Figure~\ref{fig:rfamgen_epochs} shows the mean and standard
deviation of bit scores for 1,000 sampled sequences from baseline and ECL-trained
RfamGen models at each epoch up to the selected checkpoint. The dynamics differ by
RNA family: for RF00169, ECL-trained models score higher throughout training, while
for other families the advantage appears later in training.

\end{document}